\documentclass[sigconf]{acmart}
\long\def\isnextchar#1#2#3{\def\tmpa{#2}\def\tmpb{#3}%
   \let\tmp=#1\futurelet\next\isnextcharA
}
\def\isnextcharA{\ifx\tmp\next\expandafter\tmpa\else\expandafter\tmpb\fi}
\usepackage{xspace}

\newcommand{\datadomain}{$X\in \mathbb{R}^n$\xspace}
\newcommand{\ydomain}{$Y\in \{0,1\}$\xspace}
\newcommand{\sensdomain}{$A\in \{0,1\}$\xspace}
\newcommand{\predictor}{$f$\xspace}
\newcommand{\predictortheta}{f_\theta\xspace}
\newcommand{\predictordomain}{$\hat{Y}\in \{0,1\}$\xspace}
\newcommand{\y}{$Y$\xspace}
\newcommand{\sens}{$A$\xspace}
\newcommand{\yhat}{$\hat{Y}$\xspace}
\newcommand{\task}{$\mathcal{T}$\xspace}
\newcommand{\dataset}{$\mathcal{D}$\xspace}
\newcommand{\taskdistribution}{$P(\mathcal{T})$\xspace}
\newcommand{\loss}{$\mathcal{L}$\xspace}

\newcommand{\fairregterm}{$\mathcal{R}$\xspace}

\newcommand{\inp}{\mathbf{x}}
\newcommand{\target}{\mathbf{y}}
\newcommand{\learner}{f}
\newcommand{\lossi}{\mathcal{L}_{\mathcal{T}_i}}

\usepackage{xspace}
\usepackage{amssymb}
\usepackage{amsmath}
\usepackage{graphicx}
\usepackage{algorithmic}
\usepackage{float}
\usepackage{subfig}
\usepackage[]{algorithm2e}
\usepackage{multirow}
\DeclareMathOperator*{\argmin}{\arg\!\min}

\usepackage[draft,inline,nomargin,index]{fixme}
\fxsetup{theme=color,mode=multiuser,inlineface=\itshape,envface=\itshape}
\FXRegisterAuthor{ds}{asv}{\colorbox{gray!10!white}{\color{black}Dylan}}
\FXRegisterAuthor{sf}{asf}{\colorbox{blue!10!white}{\color{black}Sorelle}}
\FXRegisterAuthor{eg}{acs}{\colorbox{red!10!white}{\color{black}Emile}}

\AtBeginDocument{%
  \providecommand\BibTeX{{%
    \normalfont B\kern-0.5em{\scshape i\kern-0.25em b}\kern-0.8em\TeX}}}

\copyrightyear{2020}
\acmYear{2020}
\setcopyright{acmlicensed}
\acmConference[FAT* '20]{Conference on Fairness, Accountability, and Transparency}{January 27--30, 2020}{Barcelona, Spain}
\acmBooktitle{Conference on Fairness, Accountability, and Transparency (FAT* '20), January 27--30, 2020, Barcelona, Spain}
\acmPrice{15.00}
\acmDOI{10.1145/3351095.3372839}
\acmISBN{978-1-4503-6936-7/20/02}

\begin{document}

% Learning to Learn Fairness
% Learning Fair Models with 
% Fairness Warnings: 
% Learning Fairly in Data Light Domains

\title{Fairness Warnings and Fair-MAML: Learning Fairly with Minimal Data}\titlenote{Partially supported by the \grantsponsor{NSF}{NSF}{} under grant \grantnum{NSF}{IIS-1633387}. 
Code can be found at: \texttt{https://github.com/dylan-slack/fairness-warnings-fair-maml}.  Thanks to Deirdre Mulligan, Charles Marx, and the other participants of the 2019 Summer Cluster on Fairness at the Simons Institute for the Theory of Computing for interesting conversations that helped to shape this work.}

\author{Dylan Slack}
\affiliation{%
  \institution{University of California, Irvine}
} \email{dslack@uci.edu}

\author{Sorelle A. Friedler}
\affiliation{%
  \institution{Haverford College}
} \email{sorelle@cs.haverford.edu}

\author{Emile Givental}
\affiliation{%
  \institution{Haverford College}}
\email{egivental@haverford.edu}

% \pagestyle{plain}

\begin{abstract}

Motivated by concerns surrounding the fairness effects of sharing and transferring fair machine learning tools, we propose two algorithms: \textit{Fairness Warnings} and \textit{Fair-MAML}.  The first is a model-agnostic algorithm that provides interpretable boundary conditions for when a fairly trained model \textit{may not} behave fairly on similar but slightly different tasks within a given domain.  The second is a fair meta-learning approach to train models that can be quickly fine-tuned to specific tasks from only a few number of sample instances while balancing fairness and accuracy.  We demonstrate experimentally the individual utility of each model using relevant baselines and provide the first experiment to our knowledge of \textit{$K$-shot fairness}, i.e. training a fair model on a new task with only $K$ data points.  Then, we illustrate the usefulness of both algorithms as a combined method for training models from a few data points on new tasks while using Fairness Warnings as interpretable boundary conditions under which the newly trained model may not be fair.

\end{abstract}


\begin{CCSXML}
<ccs2012>
<concept>
<concept_id>10010147.10010257</concept_id>
<concept_desc>Computing methodologies~Machine learning</concept_desc>
<concept_significance>500</concept_significance>
</concept>
</ccs2012>
\end{CCSXML}

\ccsdesc[500]{Computing methodologies~Machine learning}
\keywords{machine learning, fairness, meta-learning, covariate shift}

\maketitle

\section{Introduction}
\label{sec:introduction}

As machine learning tools become more responsible for decision making in sensitive domains such as credit, employment, and criminal justice, developing methods that are both fair and accurate become critical to the success of such tools.  

Correspondingly, there has been an increasing amount of academic interest in the field of fair machine learning (for surveys, see \cite{Chouldechova2018TheFO, romei2014multidisciplinary, zliobaite2015survey, barocas2018fairness}).  Research on fairness is often concerned with identifying a notion of fairness, developing an approach that mitigates the notion of fairness, and applying the approach to a variety of data sets in a supervised learning setting (see, e.g., \cite{feldman2015certifying, Hardt2016Equality, zafar2017Afairness, zafar2017Bfairness}).

However, we ask where this leaves fairness-concerned practitioners who are interested in using fair tools for their particular applications but have access to minimal or no training data. In particular, we introduce the following questions:

\begin{itemize}
    \item When can a practitioner rule out the use of a fair tool trained in a similar but slightly different context?
    \item How can a practitioner who has access to only a few labeled training points successfully train a fair machine learning model?
\end{itemize}

We suggest the relevance of these questions through the motivating scenario of recidivism prediction.  There have been calls for and extensive action towards the proliferation of criminal risk assessment tools in the United States \cite{assessingriskstevenson}.  However, there is often a disconnect between the intended use of these tools and how they are used in practice, which can lead to undesirable or ineffective results \cite{assessingriskstevenson, comparingwebjournalism}.  The LJAF, a foundation that focuses on addressing societal issues through data driven approaches, argues for a risk assessment tool that ``[can] be adopted by judges and jurisdictions anywhere in America'' and has released such a tool that has been widely used \cite{nationalmodelLJAF,assessingriskstevenson}.  We observe that minor demographic differences in the distribution of data can lead to broad effects on statistical notions of group fairness on fairly trained machine learning models.  These results could impact the ways in which fair risk prediction tools are used because such results suggest the proliferation and transfer of such methods between precincts can lead to their unreliability.  

The most related work to the proposed questions is fairness applied to transfer learning and the covariate shift problem in machine learning.  Covariate shift deals with situations where the distribution of data in application differs from the distribution of data in training.  Covariate shift is a well studied field, and there are numerous methods that attempt to train supervised learning classifiers that are robust to test distribution shifts with respect to accuracy \cite{Bickel2009DiscriminativeLU,Subbaswamy2018PreventingFD, Lipton2018DetectingAC}.  Related methods have been developed to address fairness in the covariate shift setting.  Kallus et. al. address the problem of systematic bias in data collection and use covariate shift methods to better compute fairness metrics under such conditions \cite{Kallus2018ResidualUI}.  Coston et. al. consider the situation where there are sensitive labels available in only the source or target domain and propose covariate shift methods to solve such problems \cite{coston2019fair}.

Additional work focuses on transferring fair machine learning models across domains. Madras et. al. propose a solution called LAFTR that uses an adversarial approach to create an encoder that can be used to generate fair representations of data sets and demonstrate the utility of the encoder for fair transfer learning \cite{madras18}.  Similarly, Schumman et. al. provide theoretical guarantees surrounding transfer fairness related to equalized odds and opportunity and suggest another adversarial approach aimed at transferring into new domains with different sensitive attributes \cite{transferfairnessschumann19}. Lan and Huan observe that the predictive accuracy of transfer learning across domains can be improved at the cost of fairness \cite{Lan2017DiscriminatoryT}.  Related to fair transfer learning, Dwork et. al. use a decoupled classifier technique to train a selection of classifiers fairly for each sensitive group in a data set \cite{dwork2018decoupled}.

We argue that our proposed questions are different than the existing work in the following ways.  While methods exist that address fairness and covariate shift, such methods do not address the problem of communicating to practitioners and policy makers what domain specific factors might cause a fairly trained model to fail to be fair in practice. 

Additionally, the problem of training fair machine learning models with very little task specific training data is relatively unstudied.  Practitioners might have access to minimal training data in one task and sufficient data from other related tasks.  This data might be minimal or skewed in terms of which sensitive attribute or label the data belongs to because of data collection issues associated with sensitive data sets like those discussed in Kallus et. al \cite{Kallus2018ResidualUI}.  Though LAFTR offers a way to transfer machine learning models between tasks, we observe it is unsuccessful in very data light situations.

In this paper, we propose two different methods to address the proposed problems.  First, we discuss the situation where a practitioner has no training data and must decide whether to use a fair machine learning tool trained in another similar but slightly different context.  We introduce \textit{Fairness Warnings} --- a model agnostic approach that provides \textit{interpretable boundary conditions} on fairness for when \textit{not} to apply a fair model in a different but related context because the model \textit{may} behave unfairly.  Fairness Warnings provide an interpretable model that indicates what distribution shifts to a data set's features may cause a fairly trained classifier to act unfairly in terms of a user specified notion of group fairness.  While the covariate shift problem setting allows for arbitrary changes to the testing distribution, we only consider mean shifts in this paper.  We discuss the limitations imposed by this problem restriction in section \ref{sec:problem_restrictions_warnings}.  

To provide intuition, if Fairness Warnings were trained on a recidivism classifier with respect to the $80\%$ rule of demographic parity \cite{feldman2015certifying, barocas2016big}, the model would provide conditions such as what mean shifts to the features \texttt{age} and \texttt{priors count} would cause the model to score demographic parity lower than $80\%$.  Because law enforcement agencies report general covariate crime information \cite{acquistionpreservation}, it is likely the case that precinct specific practitioners have access to these high level details and can effectively use fairness warnings to assess whether unfairly trained machine learning model may behave fairly in their application. 

Second, we consider a better route to transfer a fair machine learning model through meta-learning.  We introduce a meta-learning approach, \textit{Fair-MAML}, to quickly train models that are both accurate and fair with respect to different notions of fairness with very minimal task specific data.  Fair-MAML is based on a meta-learning algorithm called \textit{Model Agnostic Meta Learning} or \textit{MAML} \cite{Finn2017ModelAgnosticMF} that has shown success in reinforcement learning and image recognition.  Fair-MAML encourages the learning of more general notions of fairness and accuracy that allow it to achieve strong results on new tasks with only minimal data available.  In this way, Fair-MAML can escape the negative inter-distributional effects of sharing a fair machine learning model by providing a model that can be quickly fine-tuned a specific task.  We connect Fairness-Warnings and Fair-MAML by applying Fairness Warnings as boundary conditions on the fine-tuned fair meta-model.

Besides offering better ways for practitioners to implement fair machine learning models, these methods also provide ways for involved parties to \textit{question} and \textit{assess} the results of fair machine learning models.  Considering recidivism prediction, the absence of success surrounding the adoption of recidivism tools in the United States has been explained in part by judges' lack of trust in algorithm recidivism tools \cite{assessingriskstevenson,comparingwebjournalism}. Indeed, such scores are sometimes given to judges without any context \cite{assessingriskstevenson}. By including Fairness Warnings in  assessments, users may be able to better understand under what conditions the algorithm will fail to be fair. This could help increase judge understanding of such tools, as well as defense attorneys' ability to challenge its results. Additionally, by fine-tuning recidivism prediction to specific precincts using Fair-MAML, users may more readily trust that such algorithms are delivering relevant predictions than if they were only trained on disparate localities.

\section{Background}
\label{sec:background}

\subsection{Fairness}
\label{subsec:fairness}

We consider a binary fair classification setting with features \datadomain, labels \ydomain, and sensitive attributes \sensdomain.  Our goal is to train a model that outputs predictions \predictordomain such that the predictions are both accurate with respect to \y and fair with respect to the groups defined by \sens.  We consider $1$ the ``positive" outcome (being labeled at low risk of recidivism) and $0$ the ``negative" outcome (being labeled high risk).  Within the sensitive attribute, one label is protected (denoted $0$) and the other unprotected (denoted $1$).  The protected group might be historically disadvantaged groups such as women or African-Americans.

There are three often-used ways to define group fairness in this setting.

The first, \emph{demographic parity} (or statistical parity \cite{dwork12}), can be formalized as:

\begin{equation}
\label{eq:demographicparity}
\frac{P(\hat{Y}=1|A=0)}{P(\hat{Y}=1|A=1)}
\end{equation}
This is also known as a lack of disparate impact \cite{feldman2015certifying, barocas2016big} or discrimination \cite{calders2010three}.  A value closer to $1$ indicates fairness.

The second group fairness definition, \emph{equalized odds} \cite{Hardt2016Equality}, 
%Equalized odds 
requires that \yhat have equal \textit{true positive rates} and \textit{false positive rates} between groups, where values closer to $1$ indicate fairness:
\begin{equation}
\label{eq:equalizedodds}
\frac{P(\hat{Y}=1|A=0,Y=y)}{P(\hat{Y}=1|A=1,Y=y)} \mbox{~~~~~} y\in\{0,1\}
\end{equation}
This is also known as error rate balance \cite{chouldechova2017fair} or disparate mistreatment \cite{zafar2017Bfairness}.
\emph{Equal opportunity} (or equal true positive rates) introduces relaxed constraints on \ref{eq:equalizedodds} and requires the equivalence to hold only on the positive outcome in \y.  As compared to equalized odds, equal opportunity often allows for increased accuracy \cite{Hardt2016Equality}.

\subsection{Meta-Learning}
\label{subsec:metalearning}
    
Meta-learning is concerned with training models such that they can be trained on new tasks using only minimal data and few training iterations within a domain.  Meta-learning can be phrased as ``learning how to learn'' because such methods are trained on a range of tasks with the goal of being able to adapt to new tasks quickly \cite{Vanschoren2019}.  Metaphorically, this can be likened to finding a base camp (meta-model) from which you can quickly ascend to multiple nearby peaks (optimized per-task models). 

In the supervised learning setting, each task \task = $\{\mathcal{D}, \mathcal{L}\}$ where \dataset is a data set containing pairs $(X,Y)$ and \loss is a loss function.  We consider a distribution over tasks \taskdistribution  which we train the meta-model to adapt to.  Supposing the meta-model is a parameterized function $\predictortheta$ with parameters $\theta$, its optimal parameters are:

\begin{equation}
    \label{eq:optimalmetaparams}
    \theta^* = \argmin_\theta \mathbb{E}_{\mathcal{T}\thicksim p(\mathcal{T})} \mathcal{L}_\mathcal{T} (\predictortheta)
\end{equation}
This states that the optimal parameters of the model are those that minimize the loss with respect to both \loss and \dataset.   Intuitively, the model parameters should be such that they are nearly optimal for a range of tasks.  Ideally, this will mean that optimizing for any new task is quick and requires minimal data.

In the meta-learning scenario used in this paper, we train $\predictortheta$ to learn a new task $\mathcal{T}\sim P(\mathcal{T})$ using $K$ examples drawn from \task.  Additionally, we assume $\predictortheta$ can be optimized through gradient descent.  During the meta-training procedure, $K$ examples are drawn from $\mathcal{T}$.  The model is trained with respect to $K$ and $\mathcal{L}$ and the test performance is evaluated with $K$ new examples.  The use of only $K$ training examples for learning a new task is often referred to as $K$-shot learning and such methods have generally been applied to image recognition and reinforcement learning \cite{Vinyals2016MatchingNF}.  Based on the test performance, $\predictortheta$ is improved.  The meta-model $\predictortheta$ is evaluated at the end of meta-training through a set of tasks that are not included in the meta-training procedure.

\section{Methods}

\subsection{Fairness Warnings}

\subsubsection{Framework}

Similar to the formalization of LIME in Ribeiro et. al. \cite{riberosingh16}, we define fairness warnings as an interpretable model $g \in G$ where $G$ is a class of interpretable models such as decision trees or logistic regression \cite{slack19}.  Further, $g$ is a function $g:\mathbb{R}^d\rightarrow \{0,1\}$ where $\mathbb{R}^d$ is a set of distribution shifts applied to the features, labels, and sensitive values of some test data set $\mathcal{D} = \{X,Y,A\}$ under which a fair model is evaluated.  We assume \predictor is fair with respect to some notion of group fairness such as equation \ref{eq:demographicparity}, and the codomain of $g$ represents whether the potential shift may result in fair classifications according to that notion of group fairness.  Additionally, we assume that group fairness can be evaluated as fair or unfair according to some binary notion of fairness success such as the $80\%$ rule of demographic parity \cite{eightyperc, feldman2015certifying, barocas2016big}.  We assume access to a function $\mathcal{U}_f : \mathcal{D} \rightarrow \{0,1\}$ that maps between a data set and whether $f$ acts fairly on that data set according to the binary notion of group fairness.

\subsubsection{Problem Restrictions}
\label{sec:problem_restrictions_warnings}

In typical covariate shift settings, the testing distribution can be changed in any number of ways --- including being drawn from an entirely different distribution altogether.  In this application, we only consider shifts to the mean of the distribution of data that is available for training.  Under this assumption there could be more complex changes to the distribution that affect the mean but are not captured by this summary statistic and that may affect fairness.  Because we only consider a subset of the possible changes to the testing distribution, Fairness Warnings only indicate what mean shifts may lead a classifier to \textit{not} be fair and do not strongly indicate fairness if no warning is issued.  Additionally, it could be the case that Fairness Warnings predict unfairness for certain mean shifts but due to other changes to the testing distribution the classifier actually behaves fairly.  Because of these challenges, Fairness Warnings are just that---warnings that there is some evidence that suggests the model may behave unfairly with respect to a notion of group fairness.

\subsubsection{SLIM}

In practice, we use \textit{Supersparse Linear Integer Models} or \textit{SLIM} as the interpretable model $g$ \cite{Ustun2015SupersparseLI}.  SLIM creates a linear perceptron that reduces the magnitudes of the coefficients, removes unnecessary coefficients, and forces the coefficients to be integers. SLIM is a highly interpretable method that is well suited to trading off between model complexity in presentation and accuracy.  SLIM has been used in sensitive applications such as risk scoring  \cite{riskscoresustun}.   It has hyperparameters $C$ and $\epsilon$.  $C$ controls the marginal accuracy a coefficient must add to stay in the model while $\epsilon$ does the same except for the magnitude of the coefficients.

\subsubsection{Fairness Warnings Algorithm}

In order to train $g$, we generate some user specified number of perturbed versions of $\mathcal{D}$ using mean shifts.  We generate shifts for numerical features by randomly sampling from a Gaussian distribution with the standard deviation of the feature and mean zero.  The number sampled is the mean shift across the feature.  To perform the shift, we simply add the number to all the values in the feature.  We assume categorical features are one-hot encoded and thus only have two binary categorical features in $\{0,1\}$.  We shift each categorical feature by assuming each feature is drawn from a binomial distribution and use the percentage of features labeled $1$ as $p$.  We shift the feature vector by drawing a new $p$ from a Gaussian distribution $p\sim \mathcal{N}(p,1)$ and randomly sample a new vector according to $p$.  If $p$ is less than $0$ or greater than $1$, we adjust $p$ to $0$ or $1$ respectively.  Doing this a user specified number of times, we create a set of shifted variations $\mathcal{D}'$ of the original $\mathcal{D}$.  

For each shifted data set, we generate a fairness label $\mathcal{F}$ using the binary notion of group fairness $\mathcal{U}_f$.  We create a data set of mean shifted data sets and their group fairness behavior with respect to $f$, $\mathcal{Z}_f=(\mathcal{D}',\mathcal{F})$.  Finally, we train $g$ on $\mathcal{Z}_f$ using $\mathcal{D}'$ as the features and $\mathcal{F}$ as the labels.  Intuitively, we train $g$ so that it learns to predict what mean shifts may result in unfairness.  Assuming $shift(;)$ is some function that computes the mean shifting scheme above, the algorithm for generating fairness warnings is given as Algorithm \ref{alg:fairnesswarnings}.

\begin{algorithm}
\caption{Fairness Warnings}
\label{alg:fairnesswarnings}
\begin{algorithmic} {
 \REQUIRE $\mathcal{D}$: data set
 \REQUIRE $\mathcal{U}_f$: fairness notion
 \REQUIRE $g$: interpretable model
 \REQUIRE $N$: number of shifts to perform

  \STATE $\mathcal{Z}_f \leftarrow [ ]$ 
 \FOR{$i=1 : N$}
    \STATE $\mathcal{D}' \leftarrow shift(\mathcal{D})$
    \STATE $\mathcal{F} \leftarrow \mathcal{U}_f(\mathcal{D}')$
    \STATE $\mathcal{Z}_f \leftarrow (\mathcal{D}',\mathcal{F}) \bigcup \mathcal{Z}_f$
  \ENDFOR
  
  \STATE $g \leftarrow$ Train $g$ with $\mathcal{Z}_f$ using $\mathcal{D}'$ as features and $\mathcal{F}$ as labels
  return $g$
  
 }
 \end{algorithmic}
\end{algorithm}

\subsection{Fair Meta-Learning}

\begin{algorithm}
\caption{Fair-MAML}
\label{alg:fairmaml}
\begin{algorithmic}
{
\REQUIRE $p(\mathcal{T})$: distribution over tasks
\REQUIRE $\alpha$, $\beta$: step size hyperparameters
\STATE randomly initialize $\theta$
\WHILE{not done}
\STATE Sample batch of tasks $\mathcal{T}_i \sim p(\mathcal{T})$
  \FORALL{$\mathcal{T}_i$}
      \STATE Sample $K$ datapoints $\mathcal{D}=\{\inp^{(j)}, \target^{(j)}, \textbf{a}^{(j)}$\} from $\mathcal{T}_i$
      \STATE Evaluate $\nabla_\theta \lossi(\learner_\theta)$ using $\mathcal{D}$ and $\lossi$ 
      \STATE Compute updated parameters: $\theta_i'=\theta-\alpha \nabla_\theta [ \lossi(  \learner_\theta ) + \gamma_{\mathcal{T}_i} \mathcal{R}_{\mathcal{T}_i}(\learner_\theta)]$
      \STATE Sample $K$ \textit{new} datapoints $\mathcal{D}_i'=\{\inp^{(j)}, \target^{(j)}, \textbf{a}^{(j)} \}$ from $\mathcal{T}_i$ to be used in the meta-update
 \ENDFOR
 \STATE Update $\theta \leftarrow \theta - \beta \nabla_\theta \sum_{\mathcal{T}_i \sim p(\mathcal{T})} [ \lossi ( \learner_{\theta_i'}) + \gamma_{\mathcal{T}_i} \mathcal{R}_{\mathcal{T}_i}(\learner_{\theta_i'})]$ using each $\mathcal{D}_i'$ 
\ENDWHILE
}
%\STATE while 
\end{algorithmic}
\end{algorithm}

\subsubsection{$K$-shot Fairness}

In order to address the problem of learning fairly from minimal data on a new task,  we introduce the notion of \textit{$K$-shot fairness}.  Given $K$ training examples, $K$-shot fairness aims to (1) quickly train a model that is both fair and accurate on a given task.   Additionally, because the relationship between fairness and accuracy is often understood as a trade-off \cite{friedler19}, an additional aim is to (2) allow tuning of such a model so that it achieves different balances between accuracy and fairness using just $K$ training points.

The language used in this paper surrounding $K$-shot learning differs slightly from the language used in typical $K$-shot learning scenarios such as image recognition.  In $K$-shot image recognition, the goal is to learn how to distinguish between $N$ different image labels using only $K$ training examples of each type.  The training set size is then $KN$ examples.  Because we assume all the tasks to be binary labeled, all of our tasks are $2$-way.  In referencing $K$-shot fairness, we will mean that we are using $K$ training examples \textit{total}---irrespective of class label, with the assumption that all tasks are $2$-way.

\subsubsection{Fair-MAML Framework}

We expand the meta learning framework from section \ref{subsec:metalearning} such that each task includes a fairness regularization term \fairregterm and fairness hyperparameter $\gamma$.  Additionally, we require that \dataset have a protected feature $A$ such that $\mathcal{D}=(X,Y,A)$.  The goal of \fairregterm is to minimize some notion of group fairness and $\gamma$ dictates the trade off between \fairregterm and \loss.  A task is defined as $\mathcal{T} = \{\mathcal{D},\mathcal{L},\mathcal{R},\gamma \}$.  We adjust equation \ref{eq:optimalmetaparams} such that the optimal parameters are now:

\begin{equation}
    \label{eq:optimalmetaparams_fairreg}
    \theta^* = \argmin_\theta \mathbb{E}_{\mathcal{T}\thicksim p(\mathcal{T})} [\mathcal{L}_\mathcal{T} (\predictortheta) + \gamma_\mathcal{T} \mathcal{R}_\mathcal{T} (\predictortheta)]
\end{equation}

In order to train a fair meta-learning model, we adapt Model-Agnostic Meta-Learning or MAML \cite{Finn2017ModelAgnosticMF} to our fair meta-learning framework and introduce \textit{Fair-MAML}.  MAML is trained by optimizing performance of $\predictortheta$ across a variety of tasks after one gradient step.  MAML is particularly well suited to easy fairness adaption because it works with any model that can be trained with gradient descent.  The core assumption of MAML is that some internal representations are better suited to transfer learning. The loss function used by MAML is effectively the loss across a batch of task losses.  Thus, the MAML learning configuration encourages learning representations that encode more general features than a traditional learning approach.  The MAML algorithm works by first sampling a batch of tasks, computing the updated parameters $\theta$ after one gradient step of training on $K$ data points sampled from each task, and finally updating $\predictortheta$ based on the performance of $\theta$ on a new sample of $K$ points.

We modify MAML to Fair-MAML by including a fairness regularization term \fairregterm in the task losses.  The algorithm for Fair-MAML is given in algorithm \ref{alg:fairmaml}. By including a regularization term, we hope to encourage MAML to learn generalizable internal representations that strike a desirable balance between accuracy and fairness.

\subsection{Fairness Regularizers}

A variety of fairness regularizers have been proposed to handle various definitions of group fairness \cite{kamishima12,Berk2017ACF, huang2019}.  
Because MAML has shown success with the use of deep neural networks \cite{Finn2017ModelAgnosticMF}, we require regularization terms compatible with neural networks.  Methods that require the model to be linear are clearly not applicable.  In addition, Fair-MAML requires that second derivatives be computed through a Hessian-vector product in order to calculate the meta-loss function which can be computationally intensive and time-consuming.  Thus, it is critical that our fairness regularization term be quick to compute in order to allow for reasonable Fair-MAML training times.

We propose two simple regularization terms aimed at achieving demographic parity and equal opportunity that are easy to implement and extremely quick to compute.   Let $\mathcal{D}_0$ denote the protected instances in $X$ and $Y$.  The demographic parity regularizer is:

\begin{equation}
    \begin{split}
        \mathcal{R}_{dp}(\predictortheta,\mathcal{D}) & = 1 - P(\hat{Y}=1|A=0) \\ 
         & \approx  1 - \frac{1}{|\mathcal{D}_0|}\sum_{x \in \mathcal{D}_0} P(f_\theta(x) = 1)
    \end{split}
   \label{eq:di_reg}
\end{equation}
This regularizer incurs a penalty if the probability that the protected group receives positive outcomes is low. Our value assumption here is that we want to adjust the likelihood of the protected class receiving a positive outcome upwards.  Namely, we do not reduce the rate at which the unprotected class receives positive outcomes and instead adjust upwards the rate at which the protected class receives positive outcomes. 

Additionally, we consider a regularizer aimed at improving equal opportunity.  Let $\mathcal{D}_0^1$ denote the instances within $X$ that are both protected and have the positive outcome in $Y$.

\begin{equation}
    \begin{split}
        \mathcal{R}_{eop}(\predictortheta,\mathcal{D}) & = 1 - P(\hat{Y}=1|A=0,Y=1) \\ 
         & \approx  1 - \frac{1}{|\mathcal{D}_0^1|}\sum_{x \in \mathcal{D}_0^1} P(f_\theta(x) = 1)
    \end{split}
   \label{eq:equal_opp_reg}
\end{equation}

We have a similar value assumption using this regularizer as the one for demographic parity. We adjust the true positive rate of the protected class upwards and do not decrease the true positive rate of the unprotected class.  In the case of recidivism prediction, our value system could be likened to the belief that it is better not to classify more non-black defendants as high likelihood for recidivism and instead classify black defendants at a lower rate of likelihood to recidivate. 

\section{Experiments}

We first demonstrate the individual utility of both Fairness Warnings and Fair-MAML.  We then show their usefulness as a combined method.

\subsection{Fairness Warnings}
\label{subsec:fairnesswarnings}

\subsubsection{COMPAS Recidivism Experiment Setup}

We initially consider applying Fairness Warnings to the COMPAS recidivism data set.  The COMPAS recidivism data set consists of data from over $10,000$ criminal defendants from  Broward County, Florida.  It includes attributes such as the sex, age, race, and priors for the defendants.  We pre-process the data set as described in Angwin et. al. \cite{compas}.  We create a binary sensitive column for whether the defendant is African-American.  We predict the ProPublica collected label of whether the defendant was rearrested within two years.

We trained a neural network as the model, $f$, to use with Fairness Warnings.  We trained two models---one regularized for demographic parity and the other equal opportunity using the regularization terms from equations \ref{eq:di_reg} and \ref{eq:equal_opp_reg} respectively.  The demographic parity regularized model scored $58\%$ accuracy and $81\%$ demographic parity on a $20\%$ test set.  The equal opportunity regularized model scored $54\%$ accuracy and $68\%$ equal opportunity using the same test set. For the demographic parity fairness warnings, we set the fairness warnings demographic parity threshold at $80\%$.  Meaning, if the classifier scored demographic parity above $80\%$, it was deemed fair.  In the equal opportunity setting, we set the threshold to $60\%$.  We generated $2,000$ perturbed data sets, $800$ of which were classified unfairly according to demographic parity. We set $\epsilon$ to $1e-3$ and $C$ to $1e-3$.  We found that $g$ was able to classify whether the shifts applied to the perturbed data sets would result in unfair group fairness behavior with $88\%$ accuracy on a $10\%$ test set.  Using the same perturbed set, the equal opportunity regularized network was found to be unfair in $550$ of the $2,000$ perturbed examples.   Using the same hyperparameters as before, $g$ was able to classify whether the shifts would result in unfairness with respect to equal opportunity with $86\%$ accuracy.  The Fairness Warnings for the COMPAS data set is given in figure \ref{fig:ccfairnesswarning}.

\subsubsection{COMPAS Recidivism Experiment Analysis}

The COMPAS Fairness Warnings both rely on \texttt{priors\_count} and \texttt{age} to determine what mean shifts to the data set may result in unfairness.  In the demographic parity warning for instance, if the mean group age applied to $f$ were to increase by $3$ years and mean priors were to remain unchanged, the fairness warning would predict unfairness because the score total would be $(0 \cdot 20) + (3 \cdot -2) < -1$.  However, in the equal opportunity case, the same shift would not yield unfairness because $(0 \cdot 24) + (3 \cdot -2) \nless -19$.  A case that would result in unfairness in the equal opportunity setting would be a decrease in mean priors count by one charge and for age to remain level, i.e. $(-1 \cdot 24) + (0 \cdot -2) < -19$.

Overall, the SLIM implementation of fairness warnings showed good ability to classify whether certain mean shifts applied to the feature values of the COMPAS data set would result in unfairness.  Because SLIM is tunable with respect to the importance threshold of features shown in the presentation of the model, the classifier only outputs $2$ of a possible $8$ feature values in both warnings.  The presentation is simple.  A practitioner would only have to perform a few arithmetic operations in order to compute the fairness warning outcome.  

Additionally, we were able to train a random forest classifier using $200$ estimators from the Scikit-learn implementation which scored $94\%$ and $89\%$ accuracy on the demographic parity and equal opportunity fairness warnings tasks respectively.  This suggests that more robust models could serve as much more accurate fairness warnings than SLIM.  Presenting a random forest of such size in a digestible way to a user would be difficult.  However, the success of the random forest to perform this task indicates that improved interpretable methods that achieve equal levels of interpretability to SLIM but higher levels of accuracy on the fairness warnings task could serve as more desirable fairness warnings.

\begin{figure*}

  \begin{tabular}{llll}
    \toprule
    Predict UNFAIR DEMOGRAPHIC PARITY if SCORE $<$ -1 &  & & \\

    Feature & Original Mean & Score (+/- per unit increase/decrease) & Total\\
    \midrule
    priors\_count & 3.2 priors & 20 points / prior & +........ \\
    age & 34.5 years & -2 points / year & +........ \\
    \midrule
    ADD POINTS FROM ROWS 1 to 2 & & SCORE &  =........ \\
    \multicolumn{4}{l}{(Warning accuracy: 88\%, true positive rate: 88\%, true negative rate: 89\%)} \\
  \bottomrule

\end{tabular}
 \vspace{.5cm}

  \begin{tabular}{llll}
    \toprule
    Predict UNFAIR EQUAL OPPORTUNITY if SCORE $<$ -19 &  & & \\

    Feature & Original Mean & Score (+/- per unit increase/decrease) & Total\\
    \midrule
    priors\_count & 3.2 priors & 24 points / prior & +........ \\
    age & 34.5 years & -2 points / year & +........ \\
    \midrule
    ADD POINTS FROM ROWS 1 to 2 & & SCORE &  =........ \\
    \multicolumn{4}{l}{(Warning accuracy: 86\%, true positive rate: 86\%, true negative rate: 86\%)} \\
  \bottomrule

\end{tabular}

\caption{The Fairness Warnings for the COMPAS Recidivism data set for both demographic parity and equal opportunity.  ``Warning accuracy'' indicates SLIM accuracy on the Fairness Warnings prediction task.  The original model is a neural network regularized for the respective notion of fairness.  This fairness warning is meant to be read as the expected mean shift \textit{away} from the original mean of the features presented in a practitioner's application.  For instance, if priors count were to decrease $1$ prior and age were to decrease $3$ years in the demographic parity case, the score would be $(-1 \cdot 20) + (-3 \cdot -2)=$ $-14$ points.   $-14$ points $<$ $-1$ point, so the warning would predict unfairness.  Critically, the fairness warning only makes a claim surrounding unfairness.  If the model predicts a score $\geq -1$, the model \textit{does not} certify fair behavior.}
  \label{fig:ccfairnesswarning}
\end{figure*}

\subsection{Fair-MAML}

\subsubsection{Synthetic Experiment Setup}
\label{sec:synthetic_experiment}

We illustrate the usefulness of Fair-MAML as opposed to a regularized pre-trained model in fair few-shot classification through a synthetic example based on Zafar et. al \cite{zafar2017Afairness}.  We generate two Gaussian distributions using the means and covariances from Zafar et. al.  The first distribution (1) is set to $p(x) = N([2; 2], [5, 1; 1, 5])$ and the second (2) is set to $p(x) = N([-2; -2], [10, 1; 1, 3])$.  During training, we simulate a variety of tasks by dividing the class labels along a line with y-intercept of $(0,0)$ and a slope randomly selected on the range $[-5,5]$.  All points above the line in terms of their $y$-coordinate receive a positive outcome while those below are negative.  Using the formulation from Zafar et. al., we create a sensitive feature by drawing from a Bernoulli distribution where the probability of the example being in the protected class is: $p(a=0)= p(x'|y=1)/(p(x'|y=1)+p(x'|y=0))$ where $x' = [cos(\phi), -sin(\phi);sin(\phi), cos(\phi))]x$.  Here, $\phi$ controls the correlation between the sensitive attribute and class labels.  The lower $\phi$, the more correlation and unfairness.  We randomly select $\phi$ from the range $[2,4,8,16]$ to simulate a variety in fairness between tasks.

In order to assess the fine-tuning capacity of Fair-MAML and the pre-trained neural network, we introduced a more difficult fine-tuning task.  During training, the two classes were separated clearly by a line. For fine-tuning, we set each of the binary class labels to a distribution.  The positive class was set to distribution (1) and the negative class was set to distribution (2).  In this scenario, a straight line cannot clearly divide the two classes.  We assigned sensitive attributes using the same strategy as above and used a $\phi$ of $4$.  Additionally, we only gave $5$ \textit{positive-outcome} examples from the \textit{protected class}.  We hoped to simulate a situation where a fair classifier is needed on a new task, but there are only a few protected examples in the positive outcome to learn from---simulating the situation where the distribution of fine-tuning task data is biased.  An example of such a scenario could be if a practitioner needed to train a new recidivism tool and had access to only a few examples of African-Americans who had previously been labeled as low risk.   

We randomly generated $100$ synthetic tasks that we cached before training.  We sampled $5$ examples from each task during meta-training, used a meta-batch size of $32$ for Fair-MAML, and performed a single epoch of optimization within the internal MAML loop. We trained Fair-MAML for $5,000$ meta-iterations.  For the pre-trained neural network, we performed a single epoch of optimization for each task.  We trained over $5,000$ batches of $32$ tasks per batch to match the training set size used by Fair-MAML.  

The loss used is the cross-entropy loss between the prediction $f(x)$ and the true value using the demographic parity regularizer from equation \ref{eq:di_reg}.  We use a neural network with two hidden layers consisting of $20$ nodes and the ReLU activation function.  We used the softmax activation function on the last layer.  When training with Fair-MAML, we used $K=5$ examples and performed one gradient step update.  We set the step size $\alpha$ to $0.3$, used the Adam optimizer to update the meta-loss with learning rate $\beta$ set to $1e-3$.  We pre-trained a baseline neural network on the same architecture as Fair-MAML.  To one-shot update the pre-trained neural network we experimented with step sizes of $[0.01,0.1,0.2,0.3]$ and ultimately found that $0.3$ yielded the best trade offs between accuracy and fairness.  Additionally, we tested $\gamma$ values during training and fine-tuning of $[0,10]$. We present an example task in figure \ref{fig:synthetic_example} using $5$ fine-tuning points from the positive outcome and protected class. When $\gamma = 0$, Fair-MAML does not incur any fairness regularization, so the model is just MAML. 

\subsubsection{Synthetic Experiment Analysis}

In the new task, there is an unseen configuration of positively labeled points. It was not possible for positively labeled points to fall below $y=0$ during training. Fair-MAML is able to perform well with respect to both fairness and accuracy on the fine-tuning task when only biased fine-tuning data is available.  The pre-trained neural network fails at performing the new task when the fine-tuning data does not come from the original distribution of data.  This example suggests that Fair-MAML has learned a more useful internal representation for both fairness and accuracy than the pre-trained neural network.  

\bgroup
\begin{figure*}

\includegraphics[width=.95\textwidth]{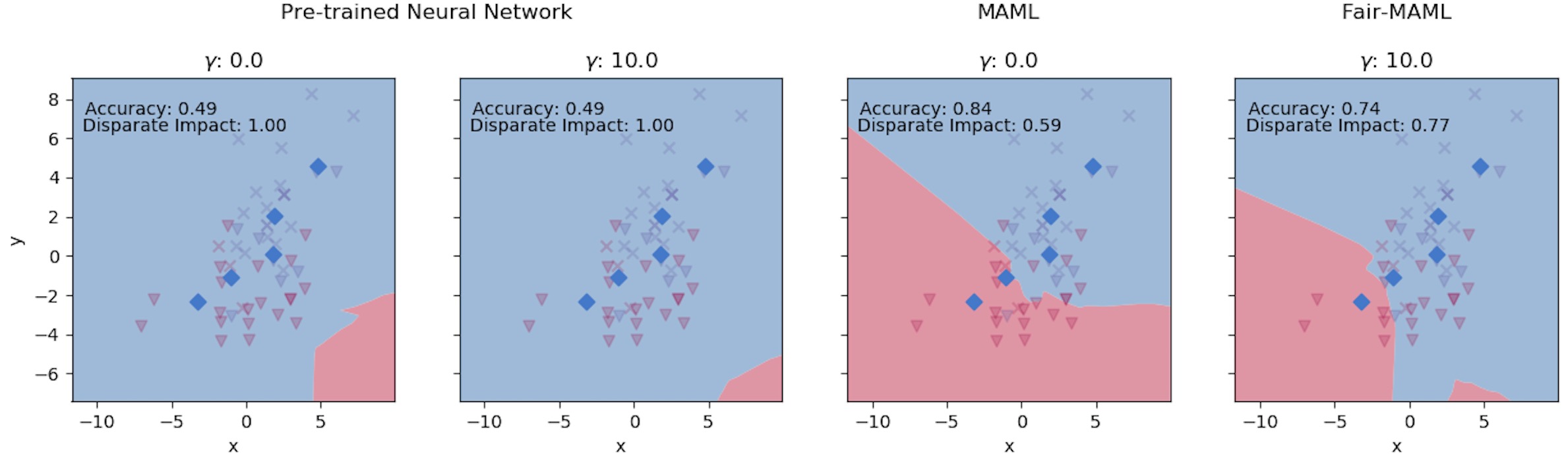}
\includegraphics[width=.95\textwidth]{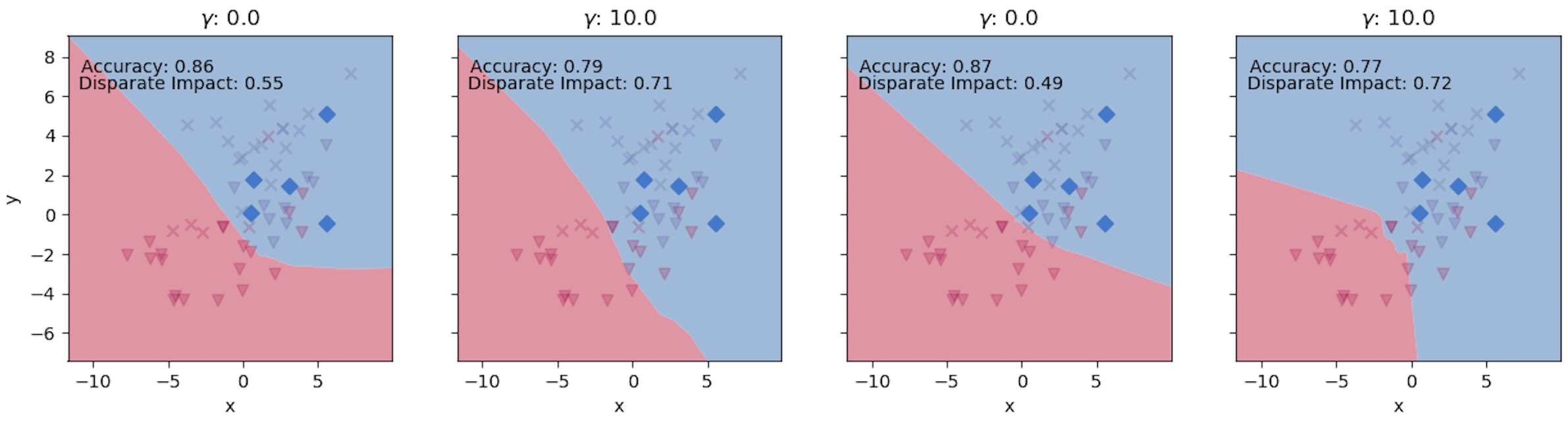}
\includegraphics[width=.95\textwidth]{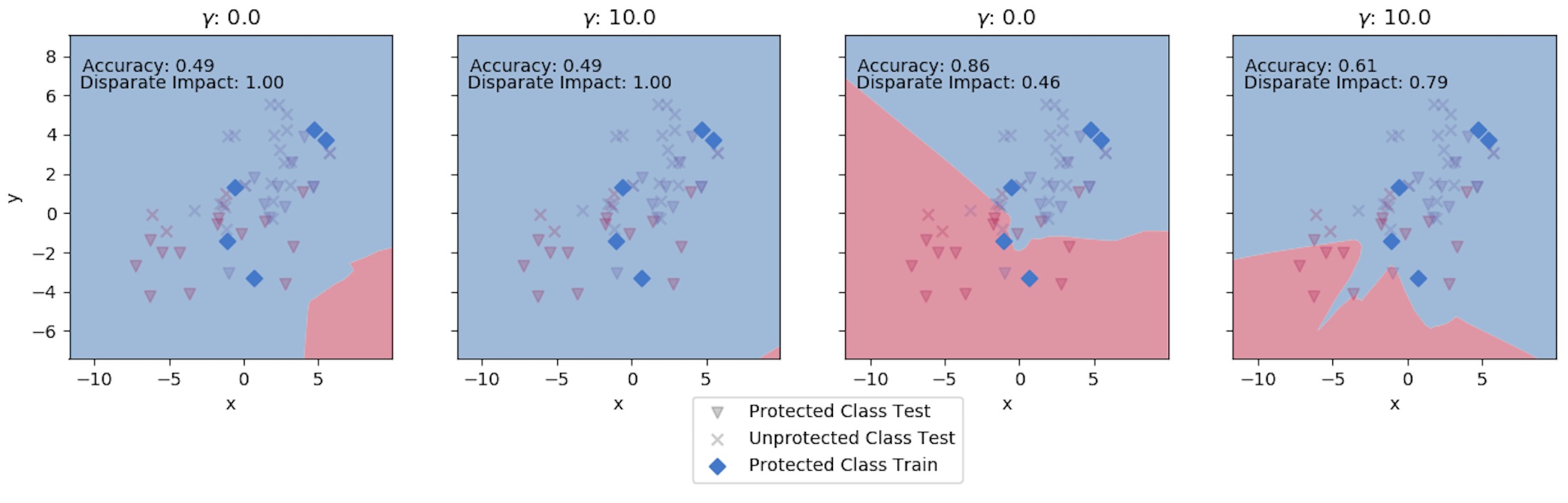}

\caption{Example decision boundaries given random draw of fine-tuning points from the pre-trained neural network, MAML, and Fair-MAML on the synthetic example (note: Fair-MAML is MAML with $\gamma=0$).  Points that are colored the same as the side of the boundary are correct.  Only points in the positive outcome and protected class are given for the fine-tuning task.  The instance where the pre-trained neural network \textit{does} provide a reasonable solution happens to come from the training distribution; otherwise, the pre-trained neural network is not able to generalize to the new situation---suggesting that Fair-MAML has learned more useful internal representations.}
\label{fig:synthetic_example}
\end{figure*}
\egroup

\subsubsection{Communities and Crime Experiment}
\label{subsubsec:ccexp}

Next we consider an example using the Communities and Crime data set \cite{communites_and_crime}.  The Communities and Crime data set includes information relevant to crime (e.g., police per population, income) as well as demographic information (such as race and sex) in different communities across the United States.  The goal is to predict the violent crime rate in the community.  We convert this data set to a few-shot fairness setting by using each state as a different task.  

Because the violent crime rate is a continuous value, we convert it into a binary label based on whether the community is in the top $50\%$ in terms of violent crime rate within a state.  Additionally, we add a binary sensitive column that receives a protected label if African-Americans are the highest or second highest population in a community in terms of percentage racial makeup.  

The Communities and Crime data set has data from $46$ states ranging in number of communities from $1$ to $278$ communities per state.  We only used states with $20$ or more communities leaving $30$ states.  We held out $5$ randomly selected states for testing and trained using $25$ states.  We set $K=10$ and cached $100$ meta-batches of size $8$ states for training.  For testing, we randomly selected $10$ communities from the hold out task that we used for fine-tuning and evaluated on whatever number of communities were left over.  The number of evaluation communities is guaranteed to be at least $10$ because we only included states with $20$ or more communities.   

We trained two Fair-MAML models---one with the demographic parity regularizer from equation \ref{eq:di_reg} and another with the equal opportunity regularizer from equation \ref{eq:equal_opp_reg}. For both models, we used a neural network with two hidden layers of $20$ nodes.  We trained the model with one gradient step using a step size of $1e-2$ and a meta-learning rate of $1e-4$ using the Adam optimizer.  We trained the model for $2,000$ meta-iterations.  

In order to assess Fair-MAML, we trained a neural network regularized for fairness using the same architecture and training data.  We fine-tuned the neural network for each of the assessment tasks.  We used a learning rate of $1e-3$ for training and assessed learning rates of $[1e-4,1e-3,1e-2,1e-1]$ for fine-tuning.  We found the fine-tuning rate of $1e-1$ to perform the best trade offs between accuracy and fairness and present results using this learning rate.  We varied $\gamma$ over $[0,4]$ incremented by $1$ for the demographic parity regularizer.  We found higher $\gamma$'s to work better for the equal opportunity regularizer and varied $\gamma$ from $[0,40]$ incremented by $10$. 

\begin{figure*}

\includegraphics[scale=.18]{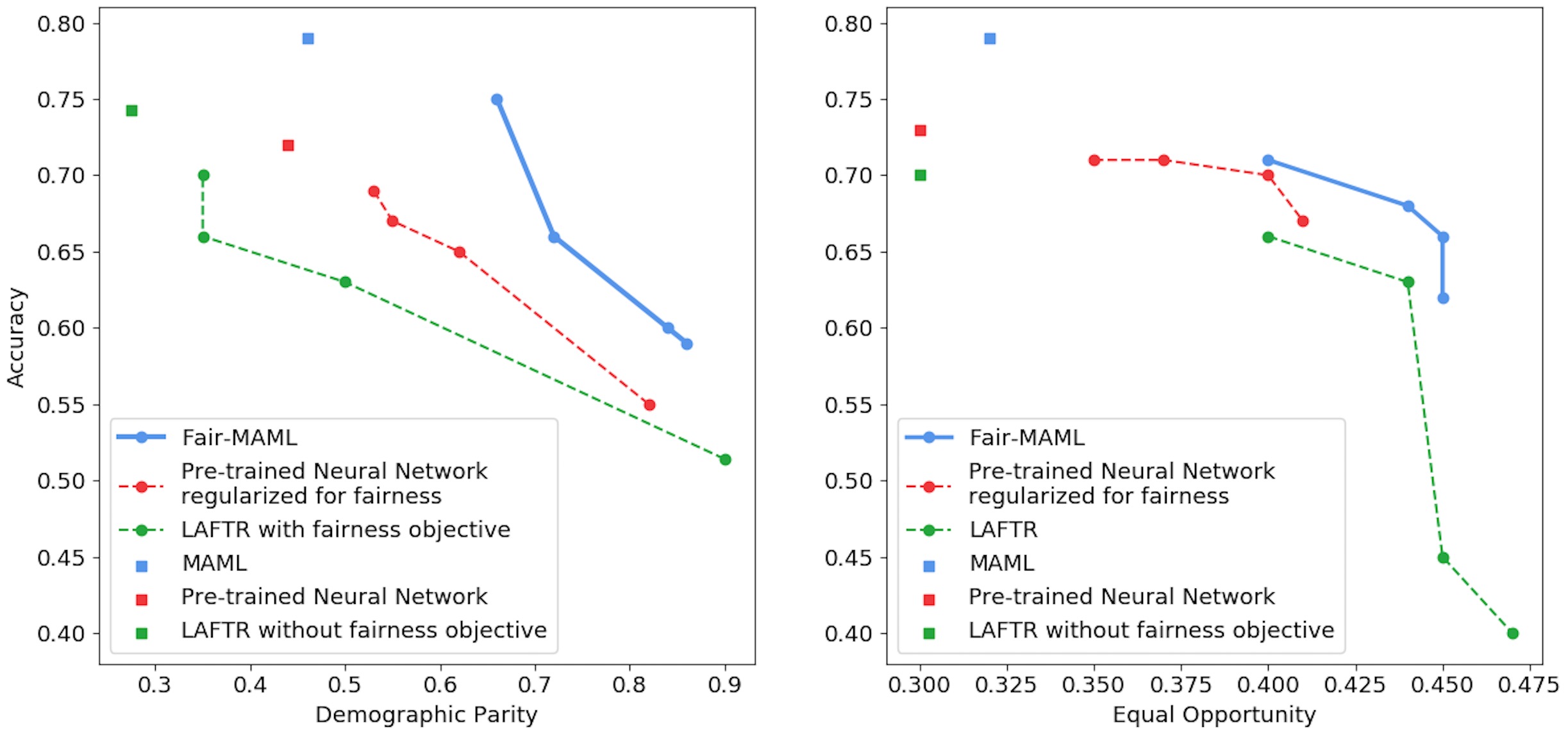}

\caption{The accuracy/fairness trade off for the communities and crimes example sweeping over a range of $\gamma$'s.  The data presented is the mean across three runs on each $\gamma$ using $5$ randomly selected hold out tasks.  The fairness numbers presented are the ratio between the protected and unprotected groups.  Higher accuracy and fairness values closer to $1.0$ indicate more successful outcomes.  The pre-trained neural network and Fair-MAML received $10$ fine-tuning points and were optimized for $1$ epoch.  We did not find useful results using LAFTR with only $10$ fine-tuning points or with a minimal number of fine-tuning epochs, so the LAFTR example given here is with $30$ fine-tuning points and $100$ epochs of optimization.  Fair-MAML is able achieve better levels of accuracy and fairness than both the pre-trained network and LAFTR on the transfer tasks using minimal fine-tuning data.}
\label{fig:cc_results}
\end{figure*}

Additionally, we trained two LAFTR models on the transfer tasks as comparisons for demographic parity and equal opportunity.  LAFTR is not intended to be compatible with our proposed $K$-shot fairness experiments because training on fine-tuning tasks with a minimal number of epochs and training points is not expected.  However, we find that it is the most relevant fair transfer learning method to use as comparison.  We used the same transfer methodology and hyperparameters as described in Madras et. al. \cite{madras18} and used a neural network with a hidden layer of $20$ nodes as the encoder.  We used another neural network with a hidden layer of $20$ nodes as the multilayer perception (MLP) to be trained on the fairly encoded representation.  We used the demographic parity and equal opportunity adversarial objectives for the first and second LAFTR model respectively.  We trained each encoder for $1,000$ epochs and swept over a range of $\gamma's$: $[0,0.5,1.0,2.0,4.0]$.  We trained with all the data not held out as one of the $5$ testing tasks. When training a MLP from the encoder on each of the transfer tasks, we found that LAFTR struggled to produce useful results with only $10$ training points from the new task over any number of training epochs.  We found that we were able to get reasonable results from LAFTR using $30$ fine-tuning points and $100$ epochs of optimization---using a minimal number of epochs was unsuccessful.  It makes sense that a minimal number of training epochs for the new task is unsuccessful because the MLP trained on the fairly encoded data is trained from scratch.  The results are presented in figure \ref{fig:cc_results}.  We were able to generate similar results with LAFTR to Fair-MAML using $50$ training points from the new task after $100$ epochs of optimization.  These results are given in the appendix. 

We observe that Fair-MAML achieves the best trade off between fairness and accuracy both in terms of demographic parity and equal opportunity.  In our proposed problem setting, LAFTR was not successful at learning with minimal data and a small number of fine-tuning epochs for the new task.  The pre-trained neural network shows some ability to learn the new task using little data and fine-tuning epochs.  At low $\gamma$'s, Fair-MAML is able to achieve higher accuracy than the pre-trained neural network and LAFTR.  Crucially, Fair-MAML is able to learn more accurate representations that are also fairer for a range of $\gamma$'s than both of the baselines.  In order to generalize to new states, only $10$ communities are needed in order to achieve strong predictive accuracy and fairness using Fair-MAML.    

\subsection{Fair-MAML with Fairness Warnings}
\label{sec:fairmamlfairnesswarnings}

\subsubsection{Motivation}
\label{subsub:combinedmamlwarningsmotivation}

We next consider Fairness Warnings applied to Fair-MAML.  We argue that Fairness Warnings can serve as a complementary tool to Fair-MAML. Because we expect Fair-MAML to be used in situations with minimal data available, it is possible that testing data given to a fine-tuned Fair-MAML model is unrepresentative of the true distribution of data for a particular task.  While in section \ref{sec:synthetic_experiment}, we empirically demonstrate that Fair-MAML can still achieve good results when training data is available from one value in a sensitive attribute or label, it still may be useful for practitioners to have indication surrounding situations in which their model may fail to be fair in testing.  

\subsubsection{Communities and Crime Fairness Warning/Fair-MAML Experiment}

We apply fairness warnings to Fair-MAML on the communities and crimes experimental setup from section \ref{subsubsec:ccexp} using demographic parity as our notion of fairness.  We randomly chose an evaluation state to apply Fairness Warnings and left the rest for meta-training.  We trained two Fair-MAML models as $f$ in fairness warnings using the demographic parity regularizer for the first model and equal opportunity regularizer for the second model.  We used $\gamma=5$ for the demographic parity Fair-MAML model and $\gamma=30$ for the equal opportunity Fair-MAML model.  We trained for $2,000$ meta-iterations in a $1$-step optimization setting, with the update learning rate set to $1e-2$ and the meta learning rate set to $1e-4$.  The demographic parity Fair-MAML model scored $87\%$ demographic parity on the test set of the fine-tuning task and accuracy of $69\%$.  The equal opportunity Fair-MAML model scored $64\%$ accuracy and equal opportunity of $63\%$.

To train Fairness Warnings on the fine-tuning task, we created $2,000$ shifted data sets of the fine-tuning test data.  We trained a Fairness Warning for both demographic parity and equal opportunity.  We used the $80\%$ rule of demographic parity in the demographic parity warning and a $60\%$ equal opportunity threshold in the equal opportunity warning.  We found that $1,034$ or close to $50\%$ of the shifted data sets were classified fairly according to $f$ with respect to demographic parity and that $1,248$ of the shifted data sets were classified fairly according to equal opportunity.  We trained SLIM using $\epsilon$ of $1e-3$ and $C$ of $1e-5$ for the demographic parity fairness warning.  We adjusted $C$ to $1e-3$ for the equal opportunity fairness warning.  

SLIM was able to predict whether the mean shifts across the features in the communities and crime data set would result in demographic parity unfairness with $71\%$ accuracy on a $10\%$ test set.  A random forest with $200$ estimators was able to predict the same task with 88\% accuracy.  In the equal opportunity setting, SLIM predicted the task with $68\%$  accuracy.  A random forest with $200$ estimators was able to perform the same task with $77\%$ accuracy.  The fairness warnings are presented in figure \ref{fig:comcrimewarnings}.

\begin{figure*}[h]

  \begin{tabular}{llll}
    \toprule
    Predict UNFAIR DEMOGRAPHIC PARITY if SCORE $<$ -3,661,000 & & & \\

    Feature & Original Mean & Score (+/- per unit increase/decrease) & Total\\
    \midrule
    mean people per family & 3.1 people & 2,000,000 points / person & +........ \\
    number of people living in urban areas & 47,700 people & -1 point / person & +........ \\
    number of people living under the poverty line & 7,590 people & -5 point / person  & +........ \\
    number of sworn full time police officers & 77.4 officers & -130,000 points / officer  & +........ \\
   
    \midrule
    ADD POINTS FROM ROWS 1 to 7 & & SCORE &  =........ \\
    
    \multicolumn{4}{l}{(Warning accuracy: 71\%, true positive rate: 72\%, true negative rate: 70\%)} \\
  \bottomrule
\end{tabular}

\vspace{.25cm}

  \begin{tabular}{llll}
    \toprule
    Predict UNFAIR EQUAL OPPORTUNITY if SCORE $<$ -2 \:\:\:\:\:\:\:\:\:\:\:\:\:\:\:\: & & & \\

    Feature & Original Mean & Score (+/- per unit increase/decrease) & Total\\
    \midrule
    police operating budget & \$3M & -2 points / \$1M & +........ \\
    \midrule
    ADD POINTS FROM ROWS 1 to 1 & & SCORE &  =........ \\
    \multicolumn{4}{l}{(Warning accuracy: 68\%, true positive rate: 68\%, true negative rate: 68\%)} \\
  \bottomrule
\end{tabular}
\caption{The Fairness Warnings for Fair-MAML applied to the communities and crime data set on the fine-tuning task.  We consider Fair-MAML trained for both demographic parity and equal opportunity.  Unlike in the COMPAS example, the features that the Fairness Warnings use are different though they both relate to aspects of policing.}
  \label{fig:comcrimewarnings}
\end{figure*}

\subsubsection{Communities and Crime Fairness Warning/Fair-MAML Analysis}

The Fairness Warning trained on the fine-tuned Fair-MAML model is able to perform reasonable prediction accuracy and generates informative results.  Particularly, it is interesting to consider that the demographic parity fine-tuned model behaves unfairly when the testing data set changes according to features such as number people living under the poverty line, in urban areas, and number of police officer.  A similar result is found in the equal opportunity setting with police operating budget.  In both the demographic parity and equal opportunity cases, the fairness warnings demonstrate that seemingly small and perhaps innocuous differences between states where Fair-MAML is trained and applied could result in unfair behavior.  For instance, the addition of a couple dozen additional police officers across communities in a state in the demographic parity case could lead to the classifier behaving unfairly.  The same is true for equal opportunity and a slight increase to the mean police operating budget. As we see in this example, reasonable real world changes to the testing distribution can result in negative changes to the group fairness of the fine-tuned Fair-MAML model.  Providing Fairness Warnings to accompany the fine-tune meta model could lend additional guidance to a practitioner and help them better understand if their model will not behave fairly in application.

\section{Limitations and Conclusions}

In this paper, we introduced Fairness Warnings and Fair-MAML.  Fairness Warnings provides an interpretable model that predicts which changes to the testing distribution will cause a model to behave unfairly.  Fair-MAML is a method that ``learns to learn" fairly and can be used to train a fair model quickly from minimal data.  We demonstrate empirically the usefulness of both methods through multiple examples on both synthetic and real data sets.  

In this work, we explore Fairness Warnings applied to mean shifts in the testing distribution.  It is a relatively straight forward extension to apply Fairness Warnings to other distribution shifts such as changes to the standard deviation.  Though we are able to generate Fairness Warnings that show useful results, they ultimately are only applied to summary statistics.  Meaning, changes to the distribution that are not captured by such statistics could affect fairness in unpredictable ways.  Thus, we only propose fairness warnings as boundary conditions under which the model \textit{may not} be fair.  In this regard, receiving a non-unfair score in fairness warnings \textit{does not} guarantee that the model will behave fairly in the new domain.  We emphasize the importance of this directionality to any lawmakers or practitioners who would be interested in using Fairness Warnings and advise that they be used only to decide against the use of certain models instead of verify that models will behave fairly.  A final limitation to our work is that we assess Fair-MAML when there are many related training tasks to learn from.  In reality, there may only be a few related training tasks available.  We leave assessing how useful Fair-MAML is on domains with only a few related training tasks to future work.

\bibliographystyle{ACM-Reference-Format}
\bibliography{bib}

\clearpage

\section*{Appendix}
\noindent\begin{minipage}{\textwidth}
\centering
    \includegraphics[scale=.40]{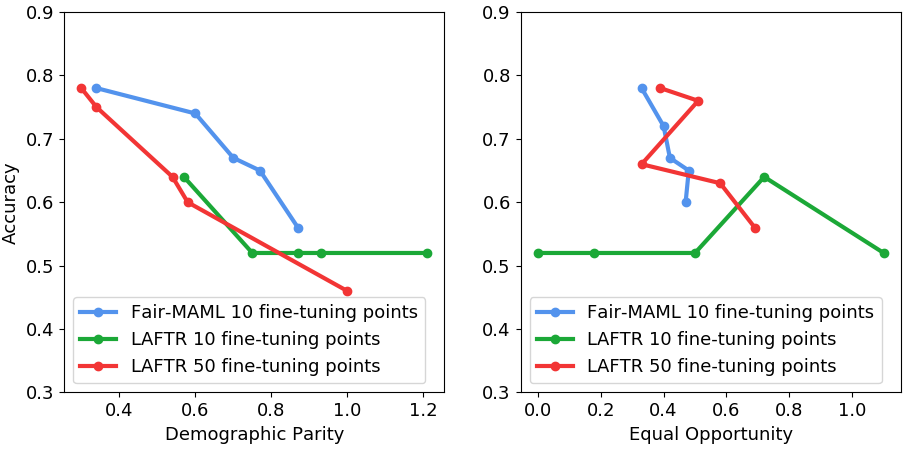}
\captionof{figure}{The accuracy/fairness trade off for the communities and crimes example sweeping over a range of $\gamma$'s considering different numbers of fine-tuning points.  The data presented is the mean across three runs on each $\gamma$ using $5$ randomly selected hold out tasks.  The states are redrawn from figure \ref{fig:cc_results} to only include those states with $60$ or more communities.  We see that with only $10$ fine-tuning points, LAFTR does not generalize well to the new tasks.  With $50$ fine-tuning points, LAFTR and Fair-MAML perform comparably.  Interestingly, Fair-MAML performs better in terms of demographic parity while LAFTR performs better in terms of equal opportunity.}
\label{fig:cc_data_efficiency}
\end{minipage}

\end{document}